\documentclass[10pt,twocolumn,letterpaper]{article}

\usepackage[pagenumbers]{cvpr} 

\usepackage{graphicx}
\usepackage{amsmath}
\usepackage{amssymb}
\usepackage{booktabs}
\usepackage{multirow}

\usepackage[pagebackref,breaklinks,colorlinks]{hyperref}

\usepackage[capitalize]{cleveref}
\crefname{section}{Sec.}{Secs.}
\Crefname{section}{Section}{Sections}
\Crefname{table}{Table}{Tables}
\crefname{table}{Tab.}{Tabs.}

\newcommand{\tumorbankDatasetName}{InUIT}

\def\cvprPaperID{7135} 
\def\confName{CVPR}
\def\confYear{2023}

\begin{document}

\title{AMIGO: Sparse Multi-Modal Graph Transformer with Shared-Context Processing for Representation Learning of Giga-pixel Images}

\author{Ramin Nakhli\\
University of British Columbia \\
{\tt\small ramin.nakhli@ubc.ca}
\and
Puria Azadi Moghadam\\
University of British Columbia
\and
Haoyang Mi \\
Johns Hopkins University
\and
Hossein Farahani \\
University of British Columbia
\and
Alexander Baras \\
Johns Hopkins University
\and
Blake Gilks \\
University of British Columbia
\and
Ali Bashashati \\
University of British Columbia \\
{\tt\small ali.bashashati@ubc.ca}
}
\maketitle

\begin{abstract}
   Processing giga-pixel whole slide histopathology images (WSI) is a computationally expensive task. Multiple instance learning (MIL) has become the conventional approach to process WSIs, in which these images are split into smaller patches for further processing.  However, MIL-based techniques ignore explicit information about the individual cells within a patch. In this paper, by defining the novel concept of shared-context processing, we designed a multi-modal Graph Transformer (AMIGO) that uses the celluar graph within the tissue to provide a single representation for a patient while taking advantage of the hierarchical structure of the tissue, enabling a dynamic focus between cell-level and tissue-level information. We benchmarked the performance of our model against multiple state-of-the-art methods in survival prediction and showed that ours can significantly outperform all of them including hierarchical Vision Transformer (ViT). More importantly, we show that our model is strongly robust to missing information to an extent that it can achieve the same performance with as low as 20\% of the data. Finally, in two different cancer datasets, we demonstrated that our model was able to stratify the patients into low-risk and high-risk groups while other state-of-the-art methods failed to achieve this goal. We also publish a large dataset of immunohistochemistry images (\tumorbankDatasetName) containing 1,600 tissue microarray (TMA) cores from 188 patients along with their survival information, making it one of the largest publicly available datasets in this context \href{https://docs.google.com/forms/d/1pRFR6j6JclZeLNsGuLKoHOvKtx1jjZc4AYGUECHl_yA/}{link to the request form}
\footnote{This paper is accepted at CVPR2023}. 
\end{abstract}

\section{Introduction}
\label{sec:intro}

Digital processing of medical images has recently attracted significant attention in computer vision communities, and the applications of deep learning models in this domain span across various image types (\eg, histopathology images, CT scans, and MRI scans) and numerous tasks (\eg, classification, segmentation, and survival prediction)~\cite{chen2022scaling,reiss2021every,xu2022closing,taleb2022contig,tang2022self,gamper2021multiple,nakhli2022ccrl}. The paradigm-shifting ability of these models to learn predictive features directly from raw images has presented exciting opportunities in medical imaging. This has especially become more important for digitized histopathology images where each data point is a multi-gigapixel image (also referred to as a Whole Slide Image or WSI). Unlike natural images, each WSI has high granularity at different levels of magnification and a size reaching 100,000$\times$100,000 pixels, posing exciting challenges in computer vision.

The typical approach to cope with the computational complexities of WSI processing is to use the Multiple Instance Learning (MIL) technique~\cite{sharma2021cluster}. More specifically, this approach divides each slide into smaller patches (\eg, 256$\times$256 pixels), passes them through a feature extractor, and represents the slide with an aggregation of these representations. This technique has shown promising results in a variety of tasks, including cancer subtype classification and survival prediction. However, it suffers from several major issues. Firstly, considering the high resolution of WSIs, even a non-overlapping 256$\times$256 window generates a huge number of patches. Therefore, the subsequent aggregation method of MIL has to perform either a simple pooling operation~\cite{ilse2018attention,boyd2021self} or a hierarchical aggregation to add more flexibility~\cite{chen2022scaling}. Nevertheless, the former limits the representative power of the aggregator drastically, and the latter requires a significant amount of computational power. Secondly, this approach is strongly dependent on the size of the dataset, which causes the over-fitting of the model in scenarios where a few data points (\eg, hundreds) are available. Lastly, despite the fact that cells are the main components of the tissue, the MIL approach primarily focuses on patches, which limits the resolution of the model to a snapshot of a population of cells rather than a single cell. Consequently, the final representation of the slide lacks the mutual interactions of individual cells.

Multiple clinical studies have strongly established that the heterogeneity of the tissue has a crucial impact on the outcome of cancer~\cite{yuan2016spatial,son2017role}. For instance, high levels of immune infiltration in the tumor stroma were shown to correlate with longer survival and positive therapy response in breast cancer patients~\cite{yuan2016spatial}. Therefore, machine learning methods for histopathology image analysis are required to account for tumor heterogeneity and cell-cell interactions. Nonetheless, the majority of the studies in 
this domain deal with a single image highlighting cell nuclei (regardless of cell type) and extra cellular matrix. Recently, few studies have investigated pathology images where various cell types were identified using different protein markers~\cite{meier2020hypothesis, wang2022cell}. However, they still utilized a single-modal approach (i.e., one cell type in an image), ignoring the multi-modal context (i.e., several cell types within the tissue) of these images.

\begin{figure}
    \centering
    \includegraphics[width=0.8\linewidth]{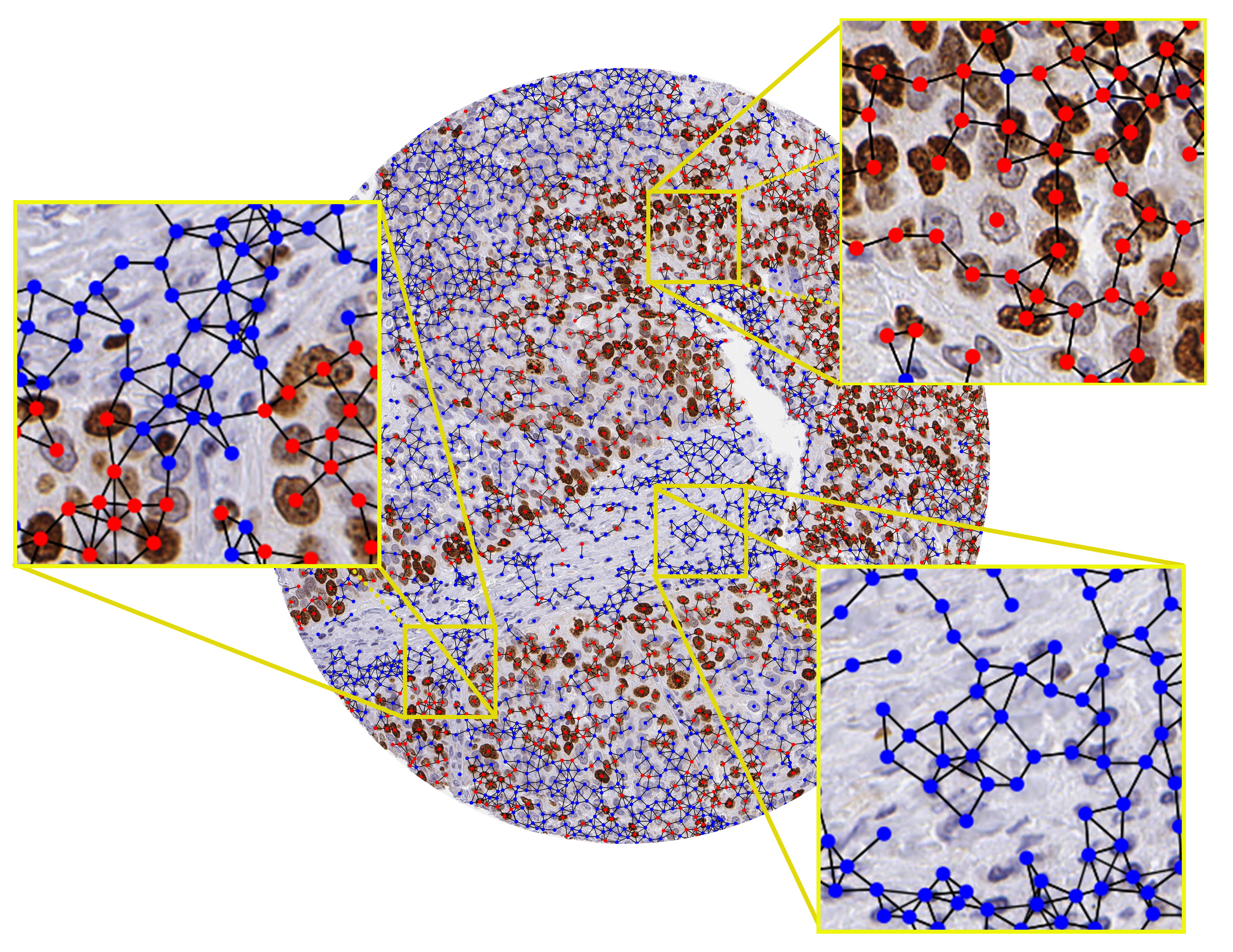}
    \caption{Cellular graph built from a \textbf{$4,000 \times 4,000$} pixel TMA core stained with Ki67 biomarker. Each red point demonstrates a cell that has a positive response to Ki67 while the blue points show cells that had a negative response to this biomarker. The highlighted patches show representative areas of the tissue where the spatial distribution of cells and the structure of the tissue are different. A typical MIL method cannot capture this heterogeneity as it does not take into account the location of the patches and lacks explicit information about the specific cells present within a patch.
    }
    \label{fig:tissue_graph}
\end{figure}

In this study, we explore the application of graph neural networks (GNN) for the processing of cellular graphs (i.e., a graph constructed by connecting adjacent cells to each other) generated from histopathology images (\cref{fig:tissue_graph}). In particular, we are interested in the cellular graph because it gives us the opportunity to focus on cell-level information as well as their mutual interactions. By delivering an adaptable focus at different scales, from cell level to tissue level, such information allows the model to have a multi-scale view of the tissue, whereas MIL models concentrate on patches with a preset resolution and optical magnification. The availability of cell types and their spatial location helps the model to find regions of the tissue that have more importance for its representation (\eg, tumor regions or immune cells infiltrating into tumor cells). In contrast to the expensive hierarchical pooling in MIL methods~\cite{chen2022scaling}, the message-passing nature of GNNs offers an efficient approach to process the vast scale of WSIs as a result of weight sharing across all the graph nodes. This approach also reduces the need for a large number of WSIs during training as the number of parameters is reduced. 

In this work, we introduce a spArse MultI-modal Graph transfOrmer model (AMIGO) for the representation learning of histopathology images by using cells as the main building blocks. Starting from the cell level, our model gradually propagates information to a larger neighborhood of cells, which inherently encodes the hierarchical structure of the tissues. More importantly, in contrast to other works, we approach this problem in a multi-modal manner, where we can get a broader understanding of the tissue structure, which is quite critical for context-based tasks such as survival prediction. In particular, for a single patient, there can be multiple histopathology images available, each highlighting cells of a certain type (by staining cells with specific protein markers), and resulting in a separate cellular graph (\cref{fig:graphical_abstract}). Therefore, using a multi-modal approach, we combine the cellular graphs of different modalities together to obtain a unified representation for a single patient. This also affirms our stance regarding the importance of cell type and the distinction between different cellular connectivity types. Aside from achieving state-of-the-art results, we notice that, surprisingly, our multi-modal approach is strongly robust to missing information, and this enables us to perform more efficient training by relying on this reconstruction ability of the network. Our work advances the frontiers of MIL, Vision Transformer (ViT), and GNNs in multiple directions:

\begin{figure*}
    \centering
    \includegraphics[width=\linewidth]{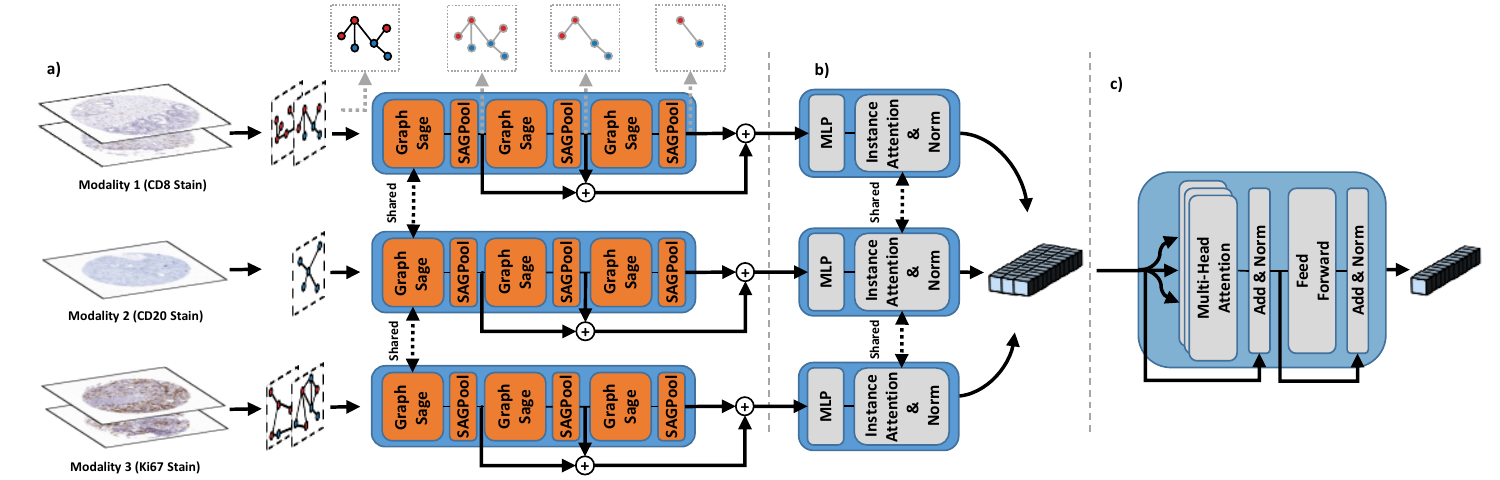}
    \caption{Overview of our proposed method. a) The Cellular graphs are first extracted from histopathology images stained with different biomarkers (\eg, CD8, CD20, and Ki67) and are fed into the encoder corresponding to their modality. The initial layer of encoders is shared, allowing further generalization, while the following layers pick up functionalities unique to each modality. The graphs at the top depict the hierarchical pooling mechanism of the model. b) The representations obtained from multiple graph instances in each modality are combined via a shared instance attention layer (shared-context processing), providing a single representation vector. c) A Transformer is used to merge the resultant vectors to create a patient-level embedding that will be used for downstream tasks such as survival prediction.}
    \label{fig:graphical_abstract}
\end{figure*}

\begin{itemize}
    \item We introduce the first multi-modal cellular graph processing model that performs survival prediction based on the multi-modal histopathology images with shared contextual information. 
    \item Our model eliminates the critical barriers of MIL models, enabling efficient training of multi-gigapixel images on a single GPU and outperforming all the baselines including ViT. It also implements the hierarchical structure of Vision Transformer while keeping the number of parameters significantly lower during end-to-end training.
    \item We also publish a large dataset of IHC images containing 1,600 tissue microarray (TMA) cores from 188 patients along with their survival information, making it one of the largest datasets in this context. 
\end{itemize}



\section{Related Work}

\subsection{Multiple Instance Learning in Histopathology} Inspired by the bag-of-words idea, Zaheer \etal ~\cite{zaheer2017deep} and Brendel \etal ~\cite{brendel2019approximating} are two pioneers of MIL models that propose the permutation-invariant bag-of-features for image representation learning. Similarly, the early works of MIL in digital pathology follow the same approach to learning representations for WSIs by relying on simple algorithms for patch-level aggregation~\cite{hou2015efficient,jia2017constrained}. However, the later works adopt more flexible designs for this purpose. For instance, IIse \etal ~\cite{ilse2018attention} use an attention-based operation to pool the representations across all patches, and Campanella \etal ~\cite{campanella2019clinical} aggregate the representation of the top-ranked patches using a Recurrent Neural Network (RNN). Li \etal ~\cite{li2021multi} introduce the idea of multi-resolution MIL by proposing a two-stage model, clustering patches at a 5$\times$ magnification and using an attention pooling on the most informative patches at the 10$\times$ magnification. Likewise, another concurrent work implements the multi-resolution idea using self-supervised learning while the authors consider the spatial positioning of the patches as well~\cite{li2021dual}. In a more recent study, Zhang \etal ~\cite{zhang2022dtfd} introduce pseudo-bagging in a double-tier setting, while Chen \etal~\cite{chen2022scaling} propose using Vision Transformers for hierarchical pooling of WSIs. Nevertheless, all the aforementioned studies ignore the cell-level details residing in the images and require a large amount of data (thousands of patients) for the training of the model. In this paper, we focus on resolving these shortcomings by using a cell-centric method while performing a hierarchical pooling of information across different sections of the image.

\subsection{Graph Neural Networks in Histopathology} Graph neural networks have recently drawn significant attention as they have led to outstanding performance in various tasks, mainly due to their structure-preserving ability~\cite{tang2021mutual,stadler2021graph}. Since this type of model works based on the foundation of local message passing, it is suitable for capturing spatial information in histopathology images~\cite{guan2022node}. 
Adnan \etal ~\cite{adnan2020representation} selected the most important patches from the WSI, created a fully-connected graph from them, and processed it using a GNN to obtain a representation for the whole graph. On the other hand, Lu \etal ~\cite{lu2020capturing} combine the adjacent similar patches into a node in the graph and then apply a GNN. In another application, Zheng \etal ~\cite{zheng2019encoding} use GNNs to provide a hashing mechanism for retrieving regions of interest that are contextually similar to the query image. Similar to our work, Chen \etal ~\cite{chen2020pathomic} and Wang \etal ~\cite{wang2022cell} use cellular graphs for survival prediction. However, unlike our proposed work, these studies ignore the type of the cell and use a single-modal setting to perform the prediction.

\subsection{Multi-Modal Image Analysis in Histopathology}
Using histopathology images along with omics data (\eg, transcriptomics and mutation) is very well studied in histopathology. For instance, Vale-Silva \etal~\cite{vale2021long} use a combination of histopathology images, clinical information, and RNA data to perform survival prediction. Chen \etal~\cite{chen2020pathomic} utilize the Kronecker product to fuse the processed histopathology image data with genomics information for survival prediction, and the same authors~\cite{chen2022pan} add more analysis to their work to link the results to interpretable features in pancreatic cancer. Although multi-modal learning of histopathology images with genomic data is extensively studied, the applications of multi-modal learning on images with different stains is vastly ignored. To the best of our knowledge, Dwivedi \etal's work ~\cite{dwivedi2022multi} is the only available study that does so to fuse different staining images for grade prediction. However, unlike our proposed design, they approach this problem in an MIL design.


\section{Method}

\subsection{Problem Formulation}

In this part, we introduce the notations used in the remainder of the paper. Consider $\{x_{n,i}^{m} | n=0, ..., N; m=0, ..., M; i=0, ..., C(n, m)\}$ to be the collection of images in a dataset, where $n$ is the patient number, $m$ is the modality number, and $i$ is the image identifier. In this setting, $N$ shows the total number of patients, $M$ is the total number of modalities, and $C(n, m)$ is the number of images available for patient $n$ from the modality of $m$. Our goal is to predict the estimated survival time of each patient, also called outcome. More specifically, we use all the available images for a patient (across different modalities) to obtain a unified representation of $R_n \in \mathbb{R}^{1\times d}$ based on which a survival time can be predicted. To avoid duplication, in the rest of this paper, we assume $x_{n,i}^{m}$ refers to both the image and the cellular graph generated from it. We will explain this pre-processing step in section \cref{section:data}.

\subsection{Multi-Modal Shared-Context Processing}
\label{section:shared_context}

\begin{figure}
  \centering
  \begin{subfigure}{0.35\linewidth}
    \includegraphics[width=\textwidth]{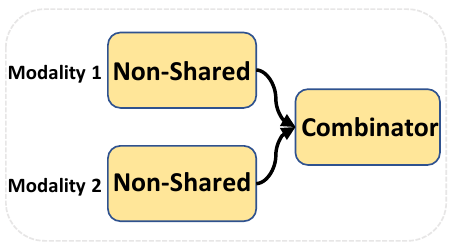}
    \caption{Non-shared-context}
    \label{fig:non_shared_context}
  \end{subfigure}
  \hfill
  \begin{subfigure}{0.63\linewidth}
    \includegraphics[width=\textwidth]{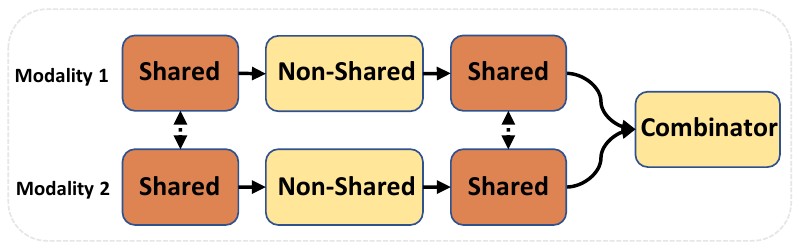}
    \caption{Shared-context}
    \label{fig:shared_context}
  \end{subfigure}
  \caption{Shared-context vs non-shared-context multi-modal processing architectures. The shared-context benefits from generalization at low-level and high-level features while allowing modality-specific processing at mid-level.}
  \label{fig:context_processing}
\end{figure}

Before delving into the specifics of our model, we must first introduce a new notion that we refer to as shared-context processing. The common strategy for processing multi-modal data is to encode each modality using a separate encoder (\cref{fig:non_shared_context}). However, we contend that combining shared and non-shared processing steps can be beneficial when dealing with different modalities containing comparable context (\eg, cellular graphs from different stains). In particular, we believe a 3-step procedure (\cref{fig:shared_context}) is necessary for such scenarios: 1) using a shared model to extract basic features from all of the modalities to help with the generalization; 2) performing modality-specific analysis using separate models for each modality; 3) unifying the high-level representations by a model shared across all modalities. This shared-context processing approach enables low-level and high-level feature unification while allowing flexibility for mid-level feature processing. In the next section, we will explain how our model benefits from this design.

\subsection{Sparse Multi-modal Cellular Graph Neural Network}

\subsubsection{Modality Encoding} In general, there are two degrees of variability for each patient: 1) a variety of images from different modalities; 2) a variety of images within each modality. The first stage of our method deals with the second type of variability and involves a processing branch specific to each modality. Each branch includes a single-modal encoder followed by an instance attention aggregator, and given $\{x_{n,i}^{m} | i=0, ..., C(n, m)\}$ as the input, it generates a single representation vector of $R_{n}^{m} \in \mathbb{R}^{1 \times d}$. 

The encoder of the branch is designed to be a GNN model consisting of three GraphSAGE layers~\cite{hamilton2017inductive}, each of which is followed by a SAGPool~\cite{lee2019self}. The SAGPooling layers enable the model to perform hierarchical pooling by selecting the most important nodes in the graph. Subsequently, the average and max pooling embeddings of the graph nodes after each SAGPool are concatenated, added together for different pooling layers, and passed through a 2-layer MLP (multi-layer perceptron).

Considering that the input graphs from different modalities have comparable context, they can benefit from the shared-context processing explained in \cref{section:shared_context}. To perform low-level feature unification, we couple the first layer of each branch using matrix factorization. More specifically, our GraphSAGE layers follow \cref{eq:coupled_graphsage}

\begin{equation}
    \hat{h}_{k}^{m} = W_{s}W_{m}[h_{k}^{m}, \frac{1}{K}\sum_{j\in\mathcal{N}_{k}}{h_{j}^{m}}].
    \label{eq:coupled_graphsage}
\end{equation}

In this equation, ${h}_{k}^{m}$ and $\hat{h}_{k}^{m}$ are the embeddings of the node $k$ before and after the layer, $W_{s}$ is the weight shared across modality branches, $W_{m}$ is the weight specific to the modality branch of $m$, $[.,.]$ is the concatenation operation, $\mathcal{N}_{k}$ is the set of nodes that are connected to node $k$, and $K$ is the size of $\mathcal{N}_{k}$. For the first GraphSAGE layer of each modality branch, we set the $W_{s}$ to be a learnable matrix, while it will be an all-ones matrix for other layers.

In order to combine the embeddings of multiple graphs of a patient within a specific modality, an instance attention, followed by an instance normalization~\cite{ulyanov2016instance}, is used similar to \cref{eq:single_modal_attention}

\begin{equation}
  R_{n}^{m} = InstanceNorm(\sum_{i=0}^{C(n, m)}{\sigma(WR_{n,i}^{m}) R_{n,i}^{m}}),
  \label{eq:single_modal_attention}
\end{equation}

where $R_{n,i}^{m}$ is the corresponding representation of $x_{n,i}^{m}$, $W$ is a learnable matrix, and $\sigma$ is the sigmoid function. The instance attention layer performs the high-level aggregation part of the shared-context processing (\cref{section:shared_context}) by sharing $W$ across all modalities.

\subsubsection{Cross-modal Aggregation}

In order to combine the representations from different modalities ($R_{n}^{m}$), we adopted a Transformer model~\cite{vaswani2017attention}. The equation of each attention head follows \cref{eq:cross_modal_attention}

\begin{equation}
    H_{n} = softmax(\frac{Q_{n} K_{n}^{T}}{\sqrt{d}})V_{n},
  \label{eq:cross_modal_attention}
\end{equation}

where $Q_{n}, K_{n}, V_{n}$ are $M \times d$ matrices resulting from applying linear transformations over the representations matrix of the patient $n$ (i.e., $concat\{R^1_n,...,R^M_n\}$). Finally, all of the heads are concatenated, passed through an MLP, and averaged across modalities to generate an embedding for the patient.

By obtaining a general understanding of the representations, the Transformer enables our model to do a cross-attention across all the stains to emphasise the most informative ones.

\subsubsection{Sparse Processing}

Despite the fact that previous studies emphasize on learning the precise topological structure of the input graph for various graph-related applications~\cite{xu2018powerful}, we find that our multi-modal approach is strongly robust to missing information. Similar to recent works in the computer vision domain (\eg, MAE~\cite{he2022masked}), we use this finding to further reduce the computational complexity of our model. More specifically, in each modality, we perform a masking operation over the feature and adjacency matrices of the input graph as shown in \cref{eq:sampling}

\begin{equation}
    \hat{X} = MX, \hat{A} = MAM^{T}, M = P\mathcal{I},
    \label{eq:sampling}
\end{equation}

where $X \in \mathbb{R}^{c \times d}$ is the feature matrix of the nodes, $A \in \mathbb{R}^{c \times c}$ is the adjacency matrix of the graph, $c$ is the number nodes in the graph, $\mathcal{I}\in \mathbb{R}^{1\times c}$ is an all-ones matrix, and $P \in \mathbb{R}^{c \times 1}$ is the mask matrix where each element comes from the Bernoulli distribution with the parameters of $1-s$, where $s$ is the sparsity ratio. As $s$ increases, the number of non-zero elements in both $\hat{X}$ and $\hat{A}$ decreases, resulting in the reduction of the subsequent computational operations. We refer to this as sparse processing and will demonstrate that our model can maintain its performance even with large sparsity ratios. 

\subsubsection{Loss Function and BCP Technique}

Survival prediction is a challenging task that includes the estimating of the failure time (death) as a continuous variable~\cite{katzman2018deepsurv}. In a maximum likelihood estimation terminology, this means that, for a subject failed at a specific time, we have to maximize the failure probability of that subject relative to the other unfailed subjects. Consider $t_j$ and $R(t_j)$ to be the time of failure for subject $j$ and the set of subjects who have survived until time $t_j$, respectively. The probability of failure for subject $j$ is calculated using \cref{eq:cox_loss}



\begin{equation}
    P_{j}(T = t_j \vert R(t_j)) = \frac{P_{j}(T = t_j \vert T \geq t_j)}{\sum_{i:t_i \geq t_j}{P_{i}(T=t_j \vert T \geq t_j)}}.
    \label{eq:cox_loss}
\end{equation}

Our training goal is to maximize this probability for each $j$. In particular, the expectation of the total loss over a mini-batch of $\mathcal{B}$ will be calculated as \cref{eq:mini_batch_loss}, in which $\mathcal{U}(.)$ is a uniform distribution over the subjects


\begin{equation}
  L_{batch} = -E_{i \sim \mathcal{U}(.)}[\log P_{i}(T = t_{i} \vert R(t_{i}))].
  \label{eq:mini_batch_loss}
\end{equation}

However, the above loss has a practical issue. The problem rises from the fact that the loss in \cref{eq:cox_loss} is only defined for subjects who have a certain time of failure, and it is undefined for subjects with a survived status in their latest follow-up (we conventionally refer to such subjects as censored data). Therefore, the censored subjects do not provide any gradient in the backpropagation step of \cref{eq:mini_batch_loss} as a separate data point due to their undefined loss (explicit gradient), which interferes with the proper training of the model. Nonetheless, one must note that such subjects still participate in the back-propagation via the denominator of \cref{eq:cox_loss} of non-censored subjects' loss (implicit gradient).

To mitigate this issue, we reformulate the loss function as \cref{eq:new_mini_batch_loss}, in which $\mathcal{U}_C(.)$ and $\mathcal{U}_N(.)$ are uniform distributions over the censored and non-censored subjects, respectively, and $k$ comes from a Bernoulli distribution with the parameter of $\alpha$ 

\begin{equation}
  L_{batch} = -E_{i \sim k \mathcal{U}_C(.) + (1-k) \mathcal{U}_N(.)}[\log P_{i}(T = t_{i} \vert R(t_{i}))].
  \label{eq:new_mini_batch_loss}
\end{equation}

One can note that for an $\alpha$ equal to the percentage of the censored cases, these two equations are equal. However, we will show that the selection of an appropriate value for this parameter results in a balanced trade-off between the implicit and explicit gradient of the censored data. We refer to $\alpha$ as batch censored portion (BCP) and show that it can have a substantial impact on the results.



\section{Experiments}

\subsection{Datasets and Pre-Processing Steps}
\label{section:data}
We used two immunohistochemistry (IHC) datasets in this study: (1) \tumorbankDatasetName: internal high-grade serous ovarian cancer cohort with 1,600 TMA cores stained with Ki67, CD8, and CD20 biomarkers collected from 188 patients, (2) MIBC: muscle-invasive bladder cancer cohort with  585 TMA cores stained with Ki67, CK20, and P16 collected from 58 patients~\cite{mi2021predictive}. Each patient has at least one TMA core stained with each biomarker, and the latest survival status (alive or dead) along with the overall survival time (since diagnosis) is available for all the patients.  

Each cell in a TMA core would appear in either a red or blue color, which shows whether the cell is positive or negative for the corresponding biomarker (\cref{fig:tissue_graph}). A cell segmentation algorithm was applied to each TMA core to locate and identify the type of cell (positive/negative)~\cite{ghahremani2021deepliif}. Then, each cell was considered as a node in the graph, and these nodes were connected to each other using a K-nearest neighbor algorithm (K=5). Similar to~\cite{javed2020cellular}, we hypothesize that there is a biological restriction on the distance of inter-cellular communications. Therefore, we removed the edges with a length of more than 60 pixels. Afterward, the embedding representation of each node was obtained by applying a pre-trained ResNet34 on a $72\times72$ crop centered on the corresponding cell in the image. Additionally, the type of node (positive/negative) and its location (relative to the size of the image) were added to the embedding as well.

\subsection{Implementation Details}

All the experiments were performed on a single GeForce RTX 3090 using Pytorch and DGL packages. The output feature size of the GraphSAGE layers and the MLP layers in each branch were set to $128$ and $32$, respectively. Adam optimizer with a learning rate of $0.002$, a cosine scheduler, a weight decay of $0.0001$, and a batch size of $128$ were used for the training of the models. The Transformer module of the cross-modal aggregator included $4$ MHSAs, and the BCP was set to $0.1$. We also used a sparsity ratio of 0.8, which was only applied at training time.

\subsection{Survival Prediction}

The summary of survival prediction results can be found in \cref{tab:survival_results}. To compare different models, similar to previous works~\cite{chen2020pathomic}, we used the concordance index (C-Index) which measures the quality of survival ranking of the patients~\cite{steck2007ranking}. Also, we used the p-value of the LogRank test to demonstrate the ability of the models in separating the high- and low-risk patients (see \cref{section:Stratification} for more information). All the experiments were performed in the 3-fold patient-wise cross-validation setting, and in contrast to the previous works, we conducted each experiment with 3 different seeds to account for the initialization variability. The results confirm that our model can outperform all of the baselines, including ViT, and has a consistent performance in both metrics across both datasets, unlike the baselines. More specifically, our model reaches the c-index of $0.57$ and $0.61$ for \tumorbankDatasetName\ and MIBC datasets while the closest baseline performances are $0.55$ and $0.59$, respectively. Additionally, our model can separate the low- and high-risk patients on both datasets significantly (p-value $< 0.01$) while being the only method to do so on the \tumorbankDatasetName\ dataset. It is worth mentioning that a few of the baseline models achieve a C-Index of $0.5$ (equivalent to random prediction), that could be attributed to the aforementioned issues of MIL-based techniques. We also notice that our model has less performance variation and number of parameters compared to the baselines, and this observation shows the generalizability and efficiency of our model, which can be linked to its cellular foundation.

The setting for the baseline models was set similar to that of ~\cite{chen2021whole}. Although the original setting used ResNet50 with a dimension of $1024$ for feature extraction, we also conducted our experiments with ResNet34 to ensure a fair comparison between our model and the baselines (more details and results in the supplementary).

\begin{table*}
  \centering
  \begin{tabular}{@{}lcccccc@{}}
    \toprule
    \multirow{2}{*}{Method}         & \multirow{2}{*}{Feature Extractor} & \multirow{2}{*}{Parameters}  & \multicolumn{2}{c}{\tumorbankDatasetName}  & \multicolumn{2}{c}{MIBC}                            \\ \cline{4-7}
                                    &                                    &                         & C-Index ($\uparrow$) & P-value ($\downarrow$) & C-Index ($\uparrow$) & P-value ($\downarrow$) \\ 
    \midrule
    \multirow{2}{*}{DeepSet}        & ResNet34           & $395K$   & $0.50 \pm 0.0$   & $0.43$  & $0.50 \pm 0.001$ & $-$                         \\ 
                                    & ResNet50           & $657K$   & $0.53 \pm 0.007$ & $0.40$  & $0.45 \pm 0.004$ & $0.28$                      \\ \cline{2-7} 
    \multirow{2}{*}{Attention MIL}  & ResNet34           & $657K$   & $0.51 \pm 0.004$ & $0.62$  & $0.59 \pm 0.007$ & $0.04$                      \\ 
                                    & ResNet50           & $920K$   & $0.55 \pm 0.004$ & $0.65$  & $0.55 \pm 0.004$ & $0.57$                      \\ \cline{2-7} 
    \multirow{2}{*}{DGC}            & ResNet34           & $658K$   & $0.53 \pm 0.007$ & $0.46$  & $0.58 \pm 0.007$ & $\mathbf{<0.001}$  \\ 
                                    & ResNet50           & $790K$   & $0.55 \pm 0.005$ & $0.31$  & $0.54 \pm 0.007$ & $0.64$                      \\ \cline{2-7} 
    \multirow{2}{*}{Patch-GCN}      & ResNet34           & $1.3M$   & $0.53 \pm 0.008$ & $0.45$  & $0.50 \pm 0.004$ & $0.005$                     \\ 
                                    & ResNet50           & $1.4M$   & $0.50 \pm 0.004$ & $0.25$  & $0.46 \pm 0.009$ & $0.33$                      \\ \cline{2-7} 
    Pathomic Fusion                 & CPC                & $368K$   & $0.51 \pm 0.001$ & $0.43$  & $0.52 \pm 0.003$ & $0.56$                      \\ \cline{2-7}
    HIPT                            & Hierarchical ViT   & $23.8M$  & $0.50 \pm 0.002$ & $0.18$  & $0.53 \pm 0.010$  & $0.1$                       \\ 
    \midrule
    AMIGO (Ours)                & ResNet34           & $451K$  & $\mathbf{0.57 \pm 0.002}$ & $\mathbf{0.01}$ & $\mathbf{0.61 \pm 0.004}$ & $\mathbf{<0.001}$ \\
    \bottomrule
  \end{tabular}
  \caption{ Survival prediction performance comparison of our model with all the baselines on two datasets. }
  \label{tab:survival_results}
\end{table*}



\subsection{Ablation Study}
\label{section:ablation}

We conducted ablation studies on different parts of our model, the results of which can be found in \cref{tab:ablation}. These experiments included the removal of the instance normalization after the instance attention (no instance norm), decoupling the modality branch weights (no weight sharing), fully coupling the modality branch weights (full weight sharing), decoupling the weights of instance attention layers (non-shared attention), no consideration of batch censored portion (no BCP), using Transformer instead of instance attention (Transformer attention), and applying sparsity at inference time (inference-time sparsity). As can be seen, depending on the dataset, each ablated feature shows a noticeable reduction in the performance of the model, emphasizing the importance of each of our design choices (more results in the supplementary).

As was elaborated, although different modalities of our data represent different stains, we believe there is a shared contextual information in all of these modalities. As a result, we can take advantage of it by employing the previously presented idea of shared-context processing. Our ablation experiments on the elimination of this step (no weight sharing and non-shared attention rows of \cref{tab:ablation}) confirm this hypothesis. On the other hand, one might argue that using the same network for all of the modalities might achieve this purpose as well. However, the corresponding ablation study (full weight sharing row of the table) invalidates this argument. Additionally, since all the modalities are processed using a shared instance attention, our model could benefit from a normalization layer before passing the embeddings to the cross-modal aggregator (no instance norm).

Our result with the removal of the BCP also demonstrates that a trade-off between the portion of the censored and non-censored data is important as it can improve the gradient signals in the backpropagation. Finally, avoiding adding sparsity at inference time results in a higher performance as the model would have access to all of the information needed for making a prediction.

\begin{table}
  \centering
  \begin{tabular}{@{}lcc@{}}
    \toprule
    \multirow{2}{*}{Ablated Feature}    & \multicolumn{2}{c}{C-Index ($\uparrow$)}           \\ \cline{2-3}
                                        & \tumorbankDatasetName                    & MIBC                       \\
    \midrule
    Baseline                            & $\mathbf{0.57 \pm 0.002}$   & $\mathbf{0.61 \pm 0.004}$   \\
    \midrule
    No instance Norm                    & $0.53 \pm 0.001$            & $0.54 \pm 0.005$            \\
    No weight sharing                   & $0.56 \pm 0.001$            & $0.54 \pm 0.016$            \\
    Full weight sharing                 & $0.53 \pm 0.001$            & $0.51 \pm 0.005$            \\
    No BCP                              & $0.54 \pm 0.001$            & $0.58 \pm 0.013$            \\
    Transformer attention               & $0.53 \pm 0.002$            & $0.55 \pm 0.011$           \\
    Inference-time sparsity             & $0.56 \pm 0.001$            & $0.55 \pm 0.004$            \\
    Non-shared attention                & $0.54 \pm 0.001$            & $0.58 \pm 0.003$            \\
    \bottomrule
  \end{tabular}
  \caption{ Ablation Studies. }
  \label{tab:ablation}
\end{table}

\subsection{Sparsity Robustness and Computational Efficiency}
\label{section:results_sparse_processing}
\begin{figure}
  \centering
  \begin{subfigure}{0.49\linewidth}
    \includegraphics[width=\textwidth]{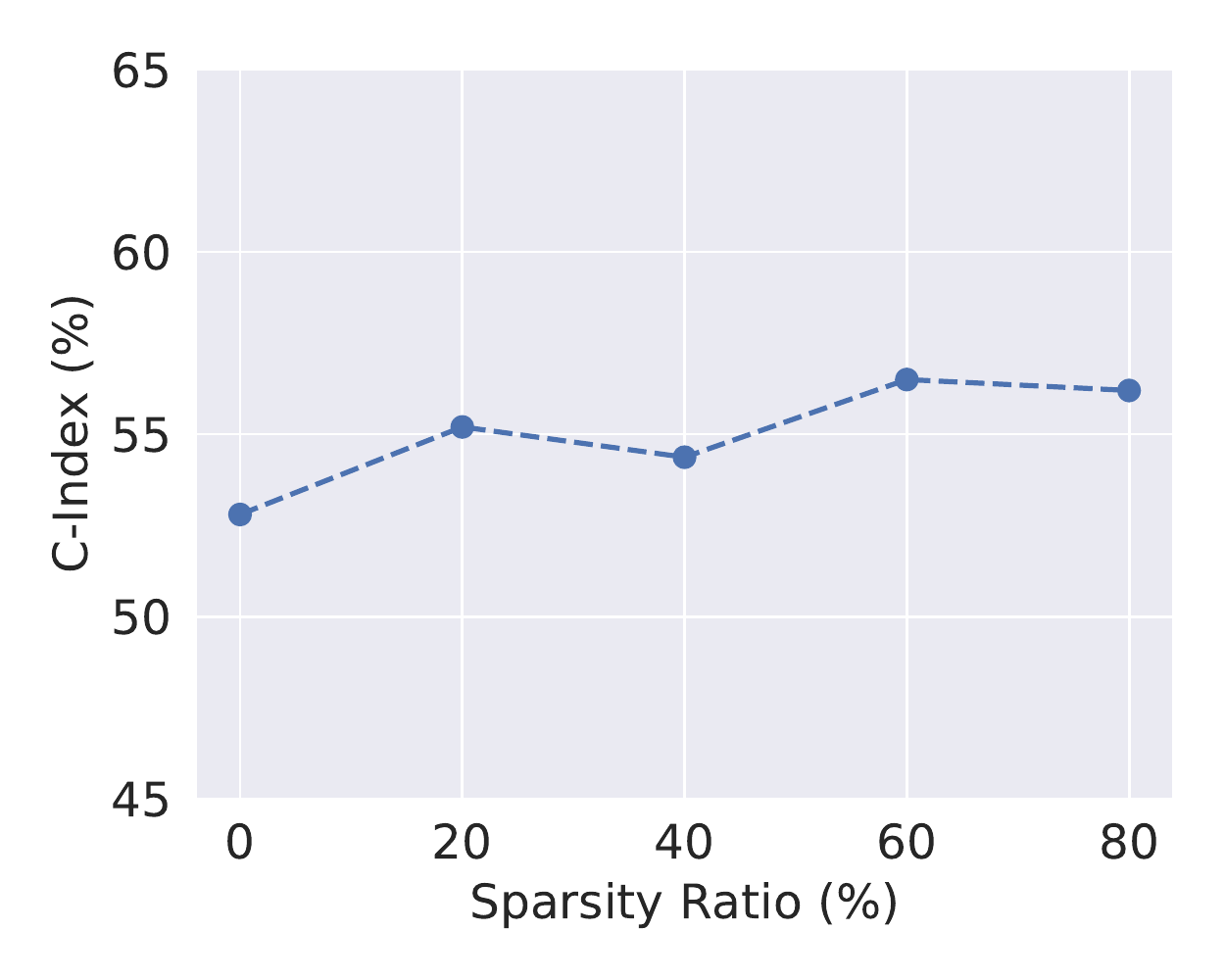}
    \caption{Sparsity C-Index}
    \label{fig:sparisty_cindex}
  \end{subfigure}
  \hfill
  \begin{subfigure}{0.49\linewidth}
    \includegraphics[width=\textwidth]{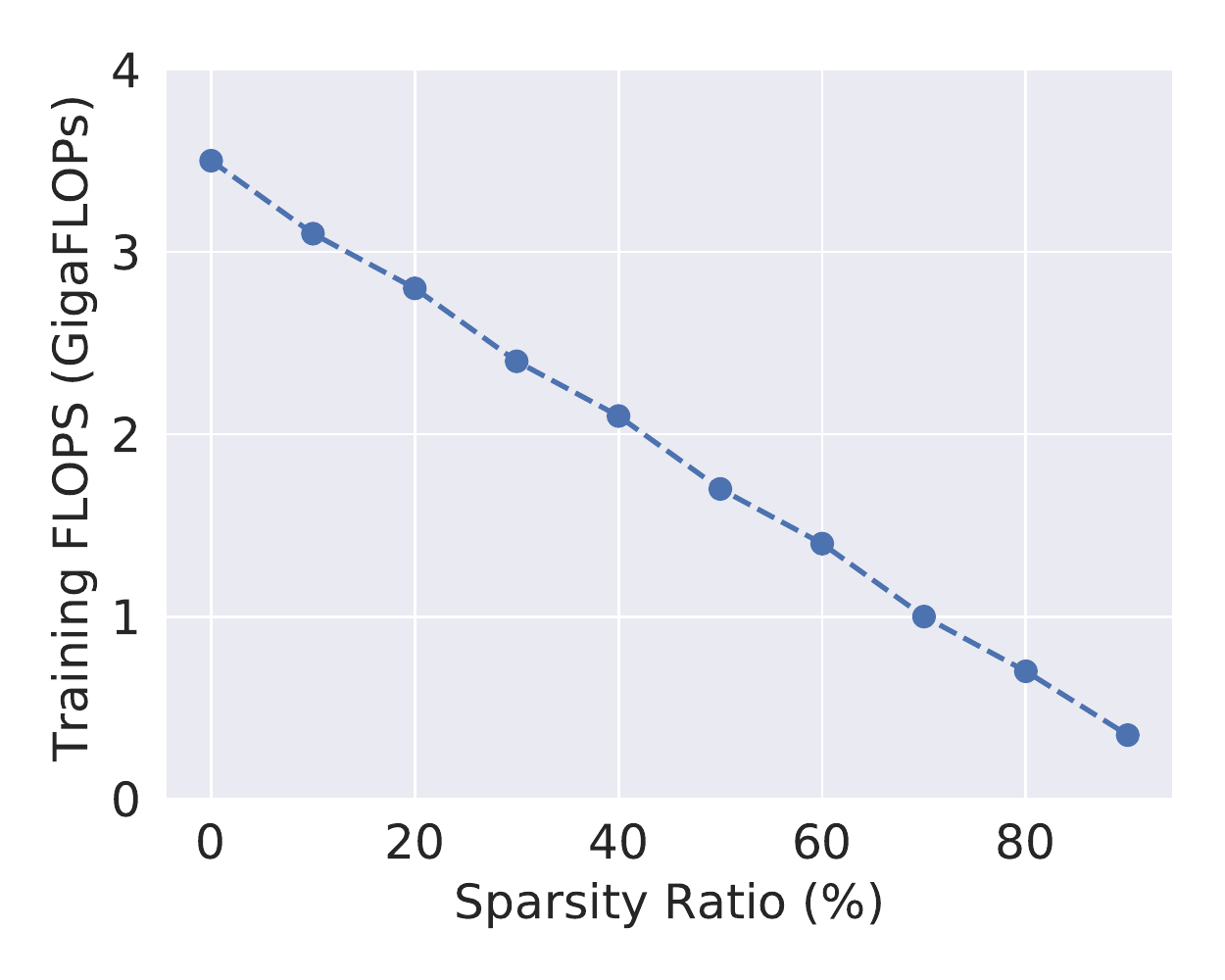}
    \caption{Sparsity Flops}
    \label{fig:sparsity_flops}
  \end{subfigure}
  \caption{\cref{fig:sparisty_cindex} shows that the final performance of our model on the \tumorbankDatasetName\ dataset is robust to the sparsity of training data. \cref{fig:sparsity_flops} demonstrates that the computational cost of our model (FLOPs) has a reverse linear relationship with the sparsity ratio. }
  \label{fig:sparsity}
\end{figure}

One of the most important findings of our study is the robustness of our model against training data sparsity. More specifically, we realized that our model's final performance is stable regardless of the sparsity ratio of the input graph. In particular, although previous digital histopathology studies~\cite{pati2022hierarchical} suggest that the learning of the complete topological structure of the cellular graph is critical for the downstream tasks, we noticed that a small sparsity ratio ($20$\%) can increase the model performance (\cref{fig:sparisty_cindex}). This observation is consistent with previous findings where they show that deep learning models can benefit from data augmentation due to the prevention of over-fitting~\cite{suresh2021adversarial}. However, the performance of our model surprisingly stays almost the same as we increase the sparsity ratio. On the other side, this sparsity ratio has a reverse linear relationship with the computational cost of the model (\cref{fig:sparsity_flops}), suggesting that higher sparsity ratios lead to a lower number of computational operations. As a result, the computational cost of our model can be significantly reduced (from 3.5 to 0.7 GigaFlops), while achieving the same performance.

\subsection{BCP Effect}
\label{section:results_bcp}
\begin{figure}
    \centering
    \includegraphics[width=0.8\linewidth]{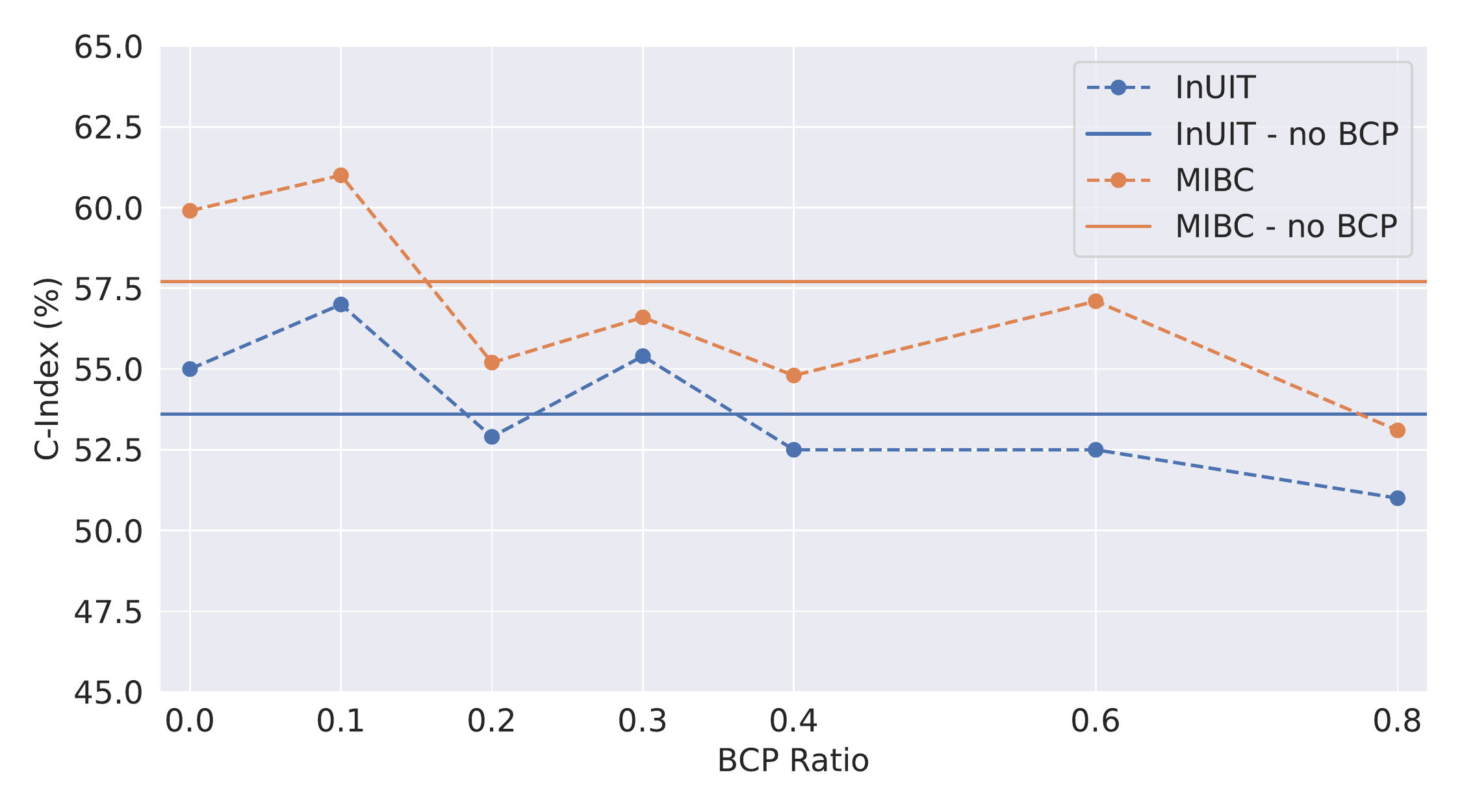}
    \caption{The performance of the model based on the ratio of BCP. Low BCP values lead to an increase in the explicit gradient and a decrease in the implicit gradient of the censored cases. A trade-off between these two types of gradients can produce the highest performance.}
    \label{fig:bcp}
\end{figure}

We also measured the effect of BCP on the performance of our model. As can be seen in \cref{fig:bcp}, a BCP of 0 (no censored data in the batch) can results in a better performance compared to the typical uniform batching as it increases the explicit gradient signals during training. On the other hand, high values of BCP result in a lower performance compared to the uniform batching as it eliminates the explicit gradient. However, the results depicted in this figure confirm our hypothesis regarding achieving the highest performance by selecting a suitable value of BCP ($0.1$) due to the trade-off between the implicit and explicit gradients. 

\subsection{Patient Stratification}
\label{section:Stratification}

\begin{figure}
  \centering
  \begin{subfigure}{0.49\linewidth}
    \includegraphics[width=\textwidth]{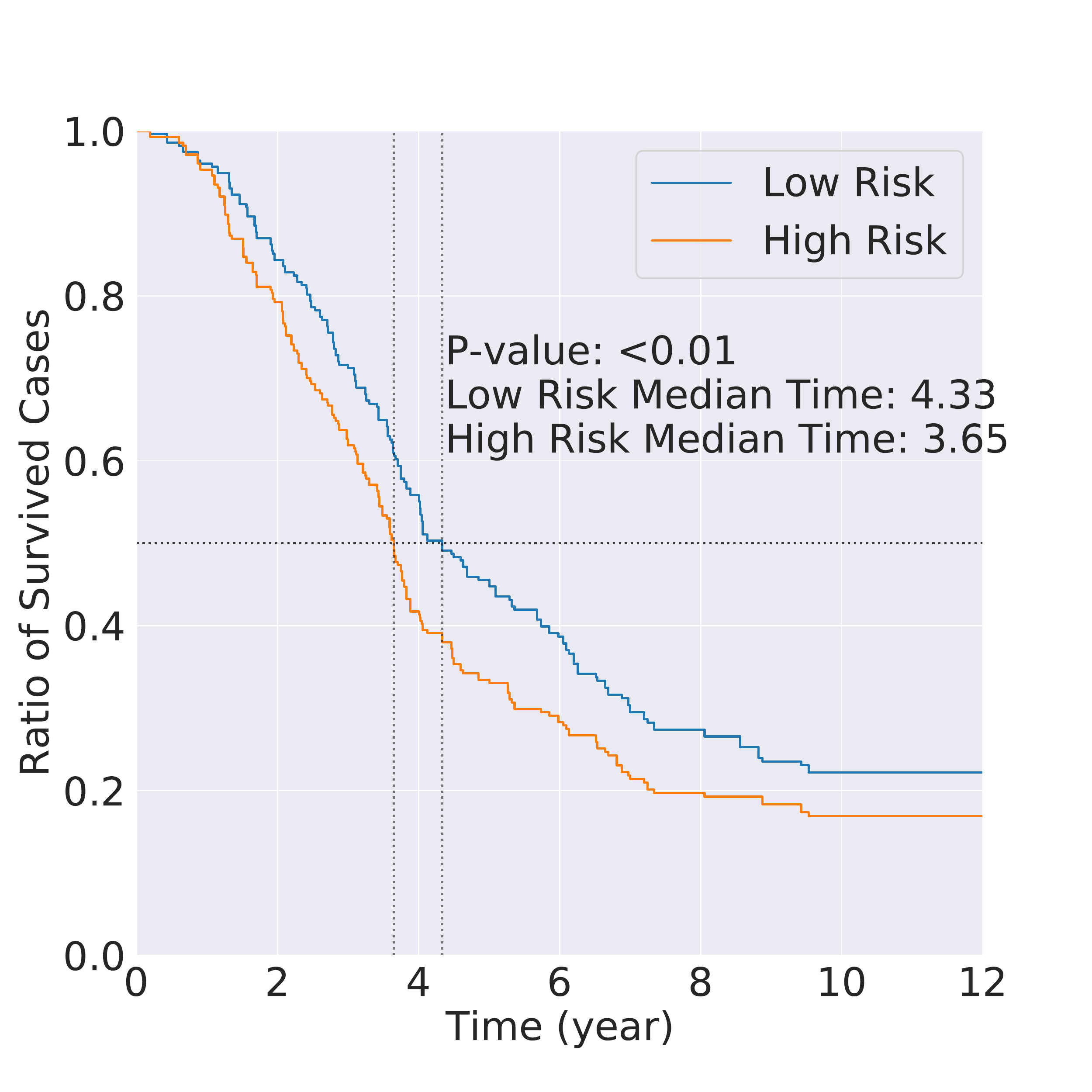}
    \caption{\tumorbankDatasetName}
    \label{fig:km_ovsurv}
  \end{subfigure}
  \hfill
  \begin{subfigure}{0.49\linewidth}
    \includegraphics[width=\textwidth]{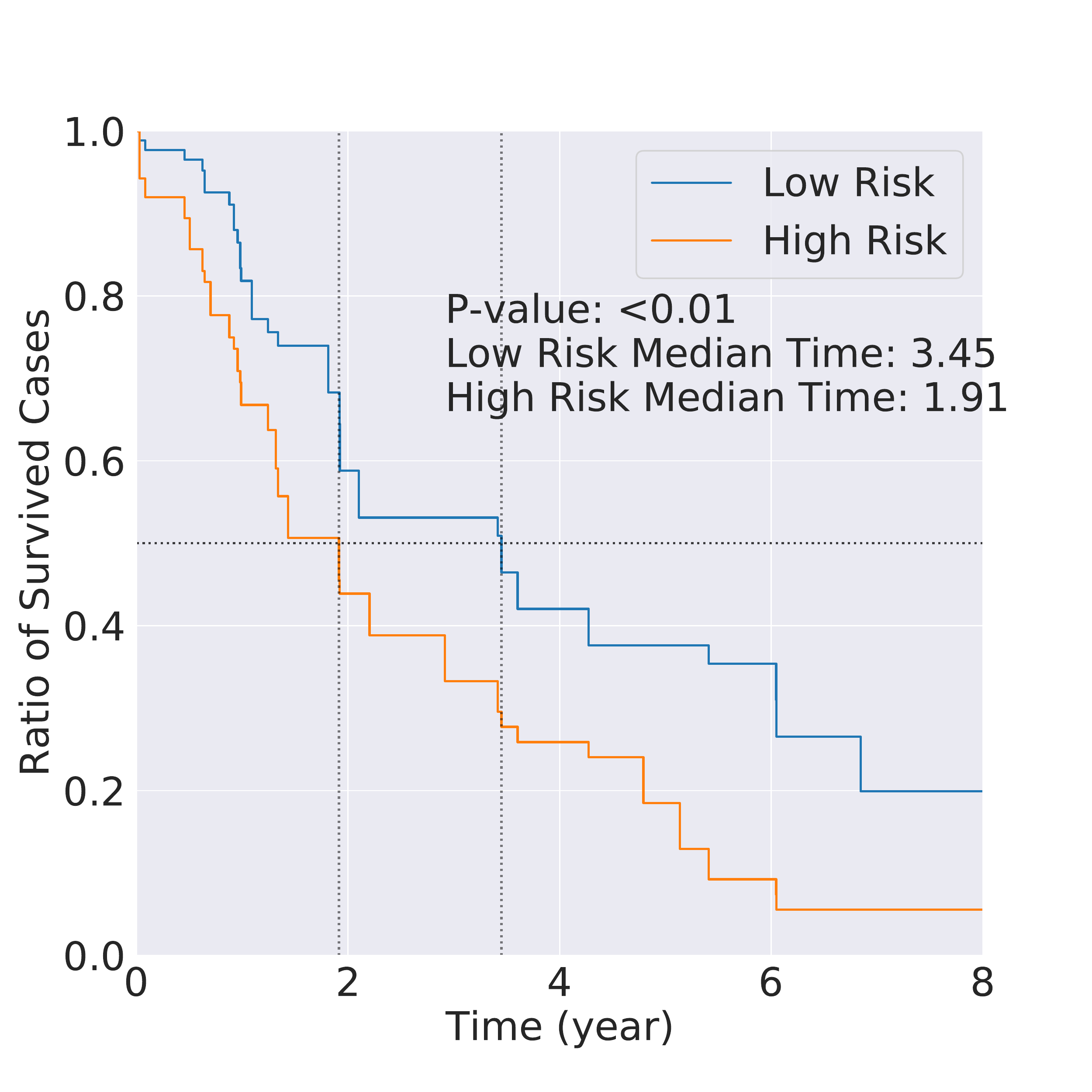}
    \caption{MIBC}
    \label{fig:km_jhu}
  \end{subfigure}
  \caption{Survival curves for cohorts of patients identified as low-risk (predicted hazard $<$ median of hazards) and high-risk (prediction hazard $>$ median of hazards) by our model.}
  \label{fig:km}
\end{figure}

While c-index is a measure that we can benchmark various survival prediction models, it is not particularly informative for patient management. For ML-based survival prediction models to become applicable in the clinic, one would need to show their utility in stratifying patients into various risk groups. Therefore, we divided the patients in each cohort into two groups (i.e., low- and high-risk) based on the predicted risk scores of our model. 
Kaplan-Meier (KM) survival curves of these cohorts are shown in \cref{fig:km}. To test the significance of the difference between the KM curves of the low- and high-risk patient categories, we employ log-rank test~\cite{bland2004logrank}.
As shown in \cref{fig:km}, our model was successful in separating both ovarian cancer and bladder cancer cases into low- and high-risk cohorts, highlighting the ability of the model in picking meaningful contextual insights from the histopathology images.
In particular, the median survival time for the high-risk and low-risk cohorts are $3.65$ and $4.33$ years for the \tumorbankDatasetName\ dataset (log-rank p-value = $0.01$) and $1.91$ and $3.45$ years for the MIBC dataset (log-rank p-value $< 0.001$), respectively. 
It is of note to mention that our findings comply with a previous study on the bladder cancer dataset, where Mi \etal~\cite{mi2021predictive} showed similar separation for the muscle-invasive bladder cancer patients. Although they used manually engineered features from the cells, we approached this problem in an end-to-end trainable manner while considering cellular interactions. 
Our ovarian cancer dataset represents a highly aggressive subtype (i.e., high-grade serous), and the majority of the efforts (though mainly unsuccessful) in the last few decades have focused on identifying biomarkers of therapy response for these patients. A study by Wang \etal~\cite{wang2017genomic} demonstrated that such markers could be found from global genomic aberration profiles and our study is the first that has led to promising results based on routine histopathology slide images. 


\section{Conclusion}

In this work, we developed, for the first time, a multi-modal GNN for the processing of histopathology images by focusing on cells and their interactions. Alongside introducing new techniques such as Batch Censored Portion (BCP) and shared-context processing, we showed that our proposed model can outperform all of its counterparts in two datasets representing ovarian and muscle-invasive bladder cancer. More importantly, we demonstrated that the proposed model is strongly robust to the sparsity of the data, to the extent that it still achieves relatively similar performance with as low as 20\% of the data during  training. By taking advantage of this observation, we were able to reduce the computational costs of the model even further. We also evaluated the applicability of our model as a tool for patient stratification, where it could split the patients into statistically significant low-risk and high-risk groups.

We believe that our proposed model highlights the importance of heterogeneity, spatial positioning, and mutual interactions of the cells for image representation across different cancer types. We hope this work can open new interesting pathways toward the efficient cell-based processing of histopathology images. Considering the success of our model in stratifying cohorts of patients that can only be separated using genomic information, we can use it as an engine to link histopathology images to gene expression, mutation, and genomic traits, where deeper analysis and biological interrogations can be performed. Furthermore, the cell-centricity of our approach offers an opportunity to identify visually-interpretable biological entities that play a key role in predicting outcomes and could be used in clinics. 

\subsubsection*{Acknowledgement}

This work was supported by the Terry Fox Research Institute, Canadian Institute of Health Research, Natural Sciences and Engineering Research Council of Canada, Michael Smith Foundation for Health Research, OVCARE Carraresi, VGH UBC Hospital Foundation, and the National Institutes of Health of United States (grant number R01CA138264).



{\small
\bibliographystyle{ieee_fullname}
\bibliography{egbib}
}

\end{document}